\documentclass[sigconf,nonacm]{acmart}
\AtBeginDocument{%
  }

\setcopyright{none}
\usepackage{xcolor}    % 提供 \textcolor
\usepackage{pifont}    % 提供 \ding

\usepackage{enumitem}
\begin{document}

% Retrieval-Augmented Generation in Retrospect and Prospect: An Information Retrieval Perspective
% \title{A Brief History of RAG: On the Past, Present, and Future}
\title{Forgotten History or Test-of-Time? Retrospect and Prospect on RAG from an IR Perspective}
% Retrospective and Prospective Perspectives on RAG from Information Retrieval
% A Retrospective and Prospective View of RAG from Information Retrieval

%%
\author{Xiaoyan Zhao}
\email{xyzhao@nus.edu.sg}
\affiliation{%
  \institution{National University of Singapore}
  \city{Singapore}
  \country{Singapore}}

\author{Yujie Cai}
\email{yujiecai@u.nus.edu}
\affiliation{%
  \institution{National University of Singapore}
  \city{Singapore}
  \country{Singapore}}

\author{Yang Zhang}
\email{zhangy@nus.edu.sg}
\affiliation{%
  \institution{National University of Singapore}
  \city{Singapore}
  \country{Singapore}}

\author{Grace Hui Yang}
\email{grace.yang@georgetown.edu}
\affiliation{%
  \institution{Georgetown University}
  \city{Washington}
  \country{United States}}

\author{Tat-Seng Chua}
\email{chuats@nus.edu.sg}
\affiliation{%
  \institution{National University of Singapore}
  \city{Singapore}
  \country{Singapore}}

\renewcommand{\shortauthors}{Zhao et al.}

\newcommand{\cmark}{\textcolor{green}{\checkmark}}
\newcommand{\xmark}{\textcolor{black}{\ding{55}}}

\newcommand{\grace}[1]{{\bf \color{red} [[Grace says ``#1'']]}}

\newcommand{\zy}[1]{\textcolor{blue}{#1}}
\newcommand{\xyz}[1]{\textcolor{blue}{[Xiaoyan: ``#1'']}}


\begin{abstract}

% Retrieval-Augmented Generation (RAG) has become a core paradigm for large language models (LLMs), enabling access to external knowledge and improving performance on knowledge-intensive tasks. While prior work primarily explains the rise of RAG in terms of the limitations of parametric LLMs, this paper revisits RAG through the lens of Information Retrieval (IR).  We argue that many key ideas underlying modern RAG and Agentic RAG—including contextual augmentation, external knowledge access, iterative refinement, and verification—have deep roots in earlier research on IR and question answering. By tracing these historical connections and mapping modern RAG components to classical IR problem formulations, we highlight substantial conceptual continuity that is often overlooked. 
% Beyond retrospective analysis, we examine why these connections have remained under-recognized and propose viewing RAG from an IR perspective by conceptualizing LLMs as a new interface layer.
% Building on this perspective, we outline how established IR principles—such as user modeling and answer validation—can more systematically inform future RAG design. By situating RAG within a broader historical trajectory of IR research, we aim to foster deeper cross-community integration, reduce unintentional rediscovery, and provide a more grounded foundation for the next generation of retrieval-augmented systems.

Retrieval-Augmented Generation (RAG) is widely regarded as a novel paradigm born from the limitations of large language models (LLMs)—a mechanism to ground their outputs in external knowledge. This view, however, is incomplete when considered within a broader historical context. In this paper, we argue that the core ideas underlying RAG are not new: foundational concepts such as integrating retrieval and language generation, knowledge augmentation, answer verification, and iterative query (or prompt) refinement had already been studied and instantiated in information retrieval (IR) and question answering (QA) research dating back to the early 2000s, well before the emergence of LLMs.

We make this case by systematically tracing the intellectual lineage of modern RAG and Agentic RAG back to their classical IR and QA antecedents, and examining why this continuity has gone under-recognized — a consequence of community fragmentation, shifting terminology, and the recency bias endemic to fast-moving fields. Rather than treating LLMs as the origin point of retrieval-augmented intelligence, we propose viewing them as a new interface layer atop a decades-old QA architecture. This reframing is not merely historical: by situating RAG within the longer trajectory of IR research, we surface underutilized prior work — on user modeling, answer validation, and query refinement — that can directly inform next-generation RAG design, reducing unintentional rediscovery and fostering genuine cross-community integration.

\end{abstract}



\maketitle

\section{Introduction}

Retrieval-Augmented Generation (RAG)~\cite{lewis2020retrieval} has become a widely adopted paradigm for Large Language Models (LLMs). RAG systems enable LLMs to ground their outputs in external knowledge and access external information during generation. This design has shown strong impact in open-domain question answering, knowledge-intensive text generation, and domain-specific applications such as biomedical analysis, legal assistance, and enterprise search~\cite{gao2023retrieval}. More recently, RAG has evolved toward Agentic RAG, where retrieval, reasoning, and generation are organized as multi-step, decision-driven processes~\cite{singh2025agentic,liang2025reasoning}. These systems allow models to iteratively issue queries, select evidence, and revise intermediate results, further expanding the scope and flexibility of generation~\cite{shao2023enhancing,asai2024self,yao2022react}.

Why has RAG emerged as a widely adopted paradigm? Most existing explanations point to the limitations of LLMs themselves. First, purely generative models are prone to hallucination and offer limited mechanisms for verification; grounding generation in retrieved documents improves factual reliability and allows outputs to be traced back to explicit sources~\cite{shuster2021retrieval,nakano2021webgpt,menick2022teaching}. Second, LLMs rely on static pretraining corpora and struggle with newly emerging or frequently updated knowledge~\cite{lewis2020retrieval,guu2020retrieval,izacard2023atlas,borgeaud2022improving}; retrieval offers a practical way to access up-to-date information without retraining the model. Third, beyond recency, many real-world applications require domain-specific or private data unavailable at pretraining time, and RAG provides a modular solution by connecting LLMs to external knowledge bases~\cite{fan2024survey,gao2023retrieval}. Some existing literature has begun connecting IR and RAG~\cite{salemi2024towards, zamani2022retrieval, kim2025towards}, though most such discussions remain confined to an LLM-centric perspective, treating retrieval primarily as a utility to augment the parametric memory of generative models.

This view, however, is incomplete when considered within a broader historical context. In this paper, we argue that the core ideas underlying RAG, including Agentic RAG, are not new: foundational concepts such as 
\begin{itemize}
    \item integrating retrieval and language generation,
    \item knowledge augmentation, 
    \item answer verification, and 
    \item iterative query (or prompt) refinement
\end{itemize}
  had already been studied and instantiated in information retrieval (IR) and question answering (QA) research dating back to the early 2000s~\cite{brill2002analysis,yang2003structured}, well before the emergence of LLMs. Long before neural language models existed, IR and QA systems were already grappling with many of the same underlying challenges that motivate RAG today: how to combine retrieved evidence with generated or synthesized answers, how to verify and rank candidate answers against source documents, and how to iteratively refine queries to improve retrieval quality. Revisiting this earlier body of work is therefore not merely a historical exercise; it offers a more principled lens for understanding what RAG is actually solving, and surfaces design patterns and lessons that current LLM-centric treatments have largely overlooked.

%To establish a deeper structural and historical alignment, we shift our perspective from the LLM itself to the broader history of IR. Viewed through this lens, RAG can be understood as integrating long-standing IR design formulations with modern generative models. The retrieve-then-generate workflow reflects a continuation of earlier efforts in generative architectures to support complex information needs through external knowledge access, structured evidence aggregation, and iterative refinement~\cite{brill2002analysis,yang2003structured} in IR. 

%This pattern can be observed across multiple stages of RAG: query reformulation in the pre-retrieval stage, matching and refinement mechanisms in retrieval, and trustworthiness and quality control in answer synthesis, which have clear counterparts in IR research. Even the recent notion of Agentic RAG, which emphasizes autonomous planning, action, and self-refinement, echoes earlier IR work on iterative retrieval and feedback-driven refinement, \textit{e.g.,} QUALIFIER~\cite{yang2003qualifier,yang2003structured,yang2002integration,yang2003trec2003qualifier}.

Specifically, the work done in TREC QA in the early 2000s, especially the QUALIFIER system, demonstrates the effectiveness of early RAG-like mechanisms across multiple stages: query reformulation in the pre-retrieval stage, matching and refinement mechanisms during retrieval, and trustworthiness and quality control in answer synthesis — each with a clear correspondence to components of current RAG systems. Even the recent notion of Agentic RAG, which emphasizes autonomous planning, action, and self-refinement, echoes this earlier IR work on iterative retrieval and feedback-driven refinement, \textit{e.g.,} QUALIFIER~\cite{yang2003qualifier,yang2003structured,yang2002integration,yang2003trec2003qualifier}.

%What distinguishes modern RAG is therefore not the emergence of entirely new design concepts, but the development of LLMs, which enable these principles to be realized at greater scale and flexibility. 

What distinguishes modern RAG, therefore, is not the emergence of entirely new design concepts, but the arrival of LLMs, which allow these long-standing IR and QA principles to be realized with far greater scale, flexibility, and generality than earlier systems could achieve. Where QUALIFIER and its contemporaries relied on hand-crafted lexical rules, heuristic patterns, and task-specific pipelines to perform query reformulation, evidence matching, and answer verification, modern RAG systems can perform these same functions implicitly, through learned representations that generalize across domains and question types with comparatively little manual engineering. In this sense, LLMs act less as a replacement for the ideas developed in early IR and QA research than as a new substrate on which those ideas can be executed more efficiently and applied far more broadly.

In this paper, we systematically trace the intellectual lineage of modern RAG back to its classical IR and QA antecedents, and suggest that this continuity has gone under-recognized due to community fragmentation, shifting terminology, and the recency bias endemic to fast-moving fields. Because this historical connection is underemphasized, many ideas reappear under new terminology as implementation tools evolve; earlier approaches are frequently rediscovered rather than explicitly reused, leading to duplicated effort across communities. Rather than treating LLMs as the origin point of retrieval-augmented intelligence, we propose viewing them as a new interface layer atop a decades-old QA architecture. This reframing is not merely historical: by situating RAG within the longer trajectory of IR research, we surface underutilized prior work — on user modeling, mixed-initiative interaction, answer validation, and query refinement~\cite{lavrenko2001relevance,carbonell1998use,thorne2018fever} — that can directly inform next-generation RAG design, reducing unintentional rediscovery and fostering genuine cross-community integration between the IR and RAG/LLM research communities.

Building on this re-connection, we present a forward-looking discussion of how RAG can more effectively leverage the accumulated efforts and insights of the IR community, identifying four important yet underdeveloped directions in today's RAG landscape. We hope this perspective contributes to the development of more principled, reliable, and user-centered retrieval-augmented systems.

\section{Proto-RAG in Early Information Retrieval}
\label{sec:rag_evolution}

RAG has rapidly become a central paradigm in the era of LLMs. 
However, long before the term “RAG” was introduced, around the turn of the century, IR research had already developed architectures that closely resemble modern retrieval–generation pipelines. This section traces these historical connections to help us better understand how these proto-RAG systems emerged and situate contemporary RAG within a broader and more coherent IR developmental trajectory.

\subsection{TREC Question Answering (QA) Track from 1999 to 2007}

A natural starting point for tracing the prehistory of RAG is the TREC Question Answering (QA) Track. The Question Answering (QA) track at the Text REtrieval Conference (TREC) ran for 9 years in its original format, from 1999 (TREC-8) through 2007. Introduced at TREC-8, the QA track provided one of the first large-scale evaluations of open-domain question answering systems \cite{voorhees2000trec8qa}. Unlike classical ad hoc retrieval, where the system returns a ranked list of documents, the QA track required systems to return short text spans that directly contained the answer to a natural-language question \cite{voorhees2001trecqa, Voorhees2002QA}. This task formulation shifted the focus of IR evaluation from document retrieval to answer finding, and it required QA systems to combine retrieval with downstream answer processing.

As a result, many TREC QA systems converged on a modular architecture consisting of question analysis and query reformulation, document or passage retrieval, candidate answer extraction, answer ranking, and answer validation \cite{uscisitrec2002, magnini2002mining, nyberg2002javelin, ChuCarrollPWCF02,ClarkeCKLLTT02,Brill2001DataIntensiveQA,lin2002extracting,yang2002integration}. Below is a pipeline commonly used by these early TREC QA systems. 
\begin{itemize}
    \item A question analysis module identifies the expected answer type and reformulates the original question into retrieval-friendly queries; 
    \item A retrieval module then narrows a large corpus to a smaller set of potentially relevant documents or passages; 
    \item Finally, an answer synthesis module extracts or generates candidate answer strings, ranks them, and verifies whether they are supported by the retrieved evidence \cite{moldovan2002performance}. 
\end{itemize}

These components closely correspond to the pre-retrieval, retrieval, and post-retrieval stages found in modern RAG systems, and the overall architecture mirrors the same retrieval-generation pipeline that underlies contemporary approaches. Despite being developed well before the term "RAG" was coined, most of these systems already treat the retrieval module as a black box: rather than modifying the internals of corpus indexing and retrieval itself, they concentrate their innovation on the techniques applied before and after the retrieval call—refining queries beforehand and processing or reranking results afterward—much in the same way modern RAG pipelines do. The one notable exception is IBM's PIQUAINT~\cite{ChuCarrollPWCF02}, which departs from this pattern by using predictive annotation to pre-annotate the corpus prior to indexing, effectively embedding retrieval-time knowledge directly into the index rather than treating retrieval as an untouched black box.

Representative systems further illustrate this connection. CMU's JAVELIN system followed a modular pipeline of question analysis, retrieval, and answer extraction, treating the retrieval stage as an interchangeable component within the larger architecture \cite{nyberg2002javelin}. The USC/ISI system took a complementary approach, building a reformulation resource that rewrote questions into multiple alternative query forms to widen retrieval coverage before extraction \cite{uscisitrec2002}. The MultiText system similarly separated passage retrieval from answer selection, applying statistical scoring methods to identify exact answers from retrieved passages \cite{ClarkeCKLLTT02}. The DIOGENE system mined recurring co-occurrence patterns from retrieved documents to support answer extraction, again relying on retrieval as an unmodified upstream step \cite{magnini2002mining}. These open-domain QA systems also incorporated question classification, named entity recognition, passage retrieval, answer selection, and external knowledge resources to improve answer extraction and verification \cite{yang2002integration,ChuCarrollPWCF02,uscisitrec2002}. 

%AskMSR reformulated each question into multiple search queries, collected Web snippets, extracted candidate answers from recurring n-grams, and ranked them using redundancy-based evidence \cite{brill2002askmsr}. This design shows that answer quality could be controlled not only through deep linguistic analysis, but also through repeated support from retrieved external evidence. 

From the perspective of RAG, these systems can be understood as proto-RAG architectures. They retrieved external evidence, selected answers conditioned on that evidence, and used ranking or validation mechanisms to control output quality. The main difference lies not in the overall problem structure, but in the implementation layer. Modern RAG replaces many hand-engineered QA modules with LLM-based query formulation, reasoning, and generation, while preserving the same core logic of evidence retrieval, answer construction, and evidence-based quality control. Below we detail the connections between these IR/QA systems and modern RAG.

%\subsubsection{Regime I: Addressing Lexical Gaps via Contextual  Augmentation}

%\subsubsection{Regime II: Addressing the Knowledge Gap via External Expansion}

\subsection{Knowledge Augmentation: Integrating Retrieval Index and Knowledge Sources}

Early QA systems were constrained by the limited coverage of closed corpora. Data sparsity made it difficult to answer open-domain questions, especially those involving rare or newly emerging entities. To overcome this knowledge gap, researchers began incorporating external resources—such as WordNet~\cite{miller1992wordnet}, the Web, knowledge bases, and later Wikipedia—into retrieval pipelines \cite{kwok2001scaling, brill2001data, brill2002analysis, zheng2002answerbus, clarke2001exploiting, radev2002probabilistic}, leveraging multiple resources to augment what could be obtained from a retrieval index built solely on a closed corpus. Representative systems include MultiText~\cite{ClarkeCKLLTT02}, MULDER~\cite{kwok2001scaling}, AskMSR~\cite{brill2001data, brill2002analysis}, and TextMap~\cite{uscisitrec2002}, which demonstrated complementary strategies for leveraging web-scale evidence. Other frameworks, such as QUALIFIER~\cite{yang2003qualifier,yang2003structured}, drew on multiple types of external resources—including WordNet and web snippets—to further enrich retrieval beyond the closed corpus.

This regime marked a structural shift: retrieval was no longer confined to a fixed local corpus but dynamically expanded to external knowledge sources. It reflects an integration of the retrieval index with other external sources, where the index and the knowledge sources can be mutually external to one another. The key insight is that neither is sufficient on its own—integration and augmentation become necessary. Modern RAG systems follow the same idea, integrating retrieval systems with LLMs, which can themselves be viewed as a huge knowledge base~\cite{petroni2019languagemodelsknowledgebases}, and for similar reasons: grounding with external evidence and expanding available knowledge. While retrieval models have evolved from lexical matching to dense and hybrid methods, the core idea—augmenting internal reasoning with external knowledge—remains consistent.

% \subsection{Pre-retrieval, Retrieval, and Post-retrieval in Methodological Context}

% In this section, we further analyze RAG techniques across pre-retrieval, retrieval, and post-retrieval stages, highlighting how they reflect/reinterpret the developmental trajectory of traditional IR.

\subsection{Query Enhancement: Retrievability, Coverage, and Control}
\label{sec:rag-preretrieval}
 Before retrieval can be effectively applied, a core challenge lies in formulating queries that better align with user intent and remain practical for real-world use. Extensive efforts have been devoted to improving this process, which can be summarized into three parts: (1) enhancing retrievability under context dependence, (2) bridging the semantic gap to improve coverage and robustness, and (3) enabling structured controllability for enforceable constraints. We present the development of each part and discuss how its technological design philosophy relates to traditional IR.

\paragraph{Part 1: Improving Retrievability via Query Rewriting} 

In the pre-retrieval stage of RAG, one focus of query enhancement is to improve retrievability, addressing a typical failure mode: a user utterance may be coherent within dialogue but lacks standalone semantic completeness, leading to information loss~\cite{ma2023query,elgohary2019can}. To mitigate this, mainstream approaches adopt context-aware query rewriting—such as pronoun resolution, ellipsis recovery, and topic completion~\cite{ram2023context,voskarides2020query}—to transform individual inputs into context-independent queries suitable for standalone retrieval.
The process closely aligns with classic IR traditions such as session-based query resolution and coreference resolution~\cite{hobbs1978resolving, lappin1994algorithm}, both of which seek to recover missing semantics from context. This reflects a core IR philosophy in RAG: normalize the query first to recover its basic meaning from context.

\paragraph{Part 2: Query Expansion}

Early retrieval-based QA systems were constrained by term-matching paradigms such as Boolean retrieval and BM25, which are sensitive to lexical mismatch between queries and documents \cite{furnas1987vocabulary, croft1995people, manning2008introduction}. When users and documents expressed the same concept using different surface forms (e.g., "laptop" vs. "notebook computer"), relevant evidence could be missed.

To address this, IR research developed query expansion (QE) techniques that enrich the original query with semantically or statistically related terms, including Rocchio-style relevance feedback \cite{rocchio1971relevance}, probabilistic relevance models such as RM3 \cite{lavrenko2001relevance}, and Local Context Analysis \cite{xu1996query}. These methods improved recall by injecting additional contextual signals while preserving the overall retrieval-based QA pipeline; as Croft \cite{croft1995people} emphasized, such expansion mechanisms became central to effective retrieval systems. Conceptually, this regime introduced contextual augmentation to address lexical mismatch—a goal modern RAG systems pursue through prompt-based reformulation, compensating for surface-level lexical limitations by incorporating externally derived context before answer construction.

Modern Agentic RAG extends this idea further: recognizing that a single query rarely captures the full latent user intent, it generates multiple retrieval probes to explore the intent space from diverse lexical and semantic angles \cite{rackauckas2024rag}, addressing the mismatch between user formulations and real-world evidence. This mirrors classic IR query expansion practices \cite{salton1990improving, xu1996query, mitra1998improving, cronen2004framework, qiu1993concept, amati2002probabilistic, zhai2001model, carpineto2012survey, diaz2016query, zamani2017relevance}, with the main differences lying in the tools used: RAG uses LLMs to generate alternative query phrasings \cite{li2024dmqr}—effectively a learned abstraction of the lexicons and thesauri traditional IR relies on \cite{miller1992wordnet, voorhees1994query}—and augments queries with generated content \cite{gao2023precise, wang2023query2doc, mao2021generation}, paralleling IR's use of pseudo-relevance labels \cite{rocchio1971relevance, lavrenko2001relevance}. RAG also enriches queries via higher-level abstraction \cite{zheng2023take} and finer-grained decomposition \cite{press2023measuring, shao2023enhancing, zhou2022least, trivedi2023interleaving}, resembling IR's relaxation/generalization \cite{salton1983extended, miller1992wordnet} and facet-based query formulation strategies \cite{yang2003structured, hearst2006clustering, kekalainen2000co}. The underlying philosophy remains consistent: expanding query representations to bridge user intent and corpus semantics.

\paragraph{Part 3: Controlled Precision via Structured Query}

This emphasis on controlled precision has an early precedent in QA research. QUALIFIER \cite{yang2003structured} constructed long, heavily structured queries that encoded constraints with a high degree of specificity, aiming for precise, well-supported answers. It followed a clear operating principle: begin with tightly constrained conditions to maximize precision, and if no answer is found, iteratively relax the constraints to allow further attempts. This relaxation was applied carefully, incrementally admitting answers into the candidate pool only as constraints loosened, so that the pool retained an appropriately high precision throughout the search. A related constrain-then-relax strategy appears in LCC \cite{Moldovan2001}, though its underlying mechanism—strict lexico-syntactic pattern matching for direct answer extraction—forgoes retrieval altogether and thus falls outside the RAG lineage this section traces.

Modern RAG systems formalize this same constrain-then-relax intuition into an explicit architectural principle, producing structured query specifications that separate soft intent for semantic matching from hard constraints for enforceable filtering, so that mandatory conditions become executable structures. There are two research lines: (1) Metadata-aware retrieval integrates metadata into query rewriting and indexing, enabling explicit constraints on section, entity, time, and document type~\cite{yousuf2026utilizing, dadopoulos2025metadata}; (2) rule-based filtering removes incompatible evidence before generation, reducing boundary violations and improving faithfulness~\cite{wang2024blendfilter, guo2025dsrag}.
These advances directly inherit from structured query studies in IR, which had already separated topical matching from constraint satisfaction through fielded queries and operator composition, treating structured operators and field constraints as first-class objects~\cite{turtle1989inference, callan1992inquery}.

%Beyond query expansion, RAG also emphasizes precision under control: natural queries often under-specify non-negotiable constraints, making retrieval hard to bound and audit. RAG therefore produces structured query specifications, separating soft intent for semantic matching from hard constraints for enforceable filtering, so that mandatory conditions become executable structures. There are two research lines: (1) Metadata-aware retrieval integrates metadata into query rewriting and indexing, enabling explicit constraints on section, entity, time, and document type~\cite{yousuf2026utilizing, dadopoulos2025metadata}; (2) Integrate rules for removing incompatible evidence before generation, reducing boundary violations and improving faithfulness~\cite{wang2024blendfilter, guo2025dsrag}. The advances directly inherit structured query IR, which has already separated topical matching from constraint satisfaction through fielded queries and operator composition. Some works treated structured operators and field constraints as first-class objects~\cite{turtle1989inference, callan1992inquery}.

\subsection{Evidence Retrieval: Matching Mechanisms}
\label{sec:rag-retrieval}

% We next focus on the retrieving module in RAG. This step concerns how queries are matched with evidence, motivating the study of matching mechanisms. %From a dynamic perspective, the key question becomes how retrieval can be iteratively refined to improve effectiveness. Accordingly, we discuss the development of matching and refinement mechanisms in RAG, highlighting their close connections to classical IR. 

% %\paragraph{Part 1: Matching Mechanism.}
% To improve retrieval precision, RAG has explored a wide spectrum of matching mechanisms.
% These range from \emph{sparse retrieval} that emphasizes lexical matching~\cite{robertson2009probabilistic,formal2021splade},
% to \emph{dense retrieval} that captures semantic similarity in vector space~\cite{karpukhin2020dense,xiong2020approximate},
% and further to \emph{graph- or tree-based retrieval} that models structured relations~\cite{sun2018open,sarthi2024raptor, edge2024local}.
% More recently, \emph{learning-based matching} functions have been introduced to align retrieval with the utility of downstream generation.
% Clearly, these developments closely mirror the evolution of traditional IR, where similar directions have been extensively studied~\cite{robertson2009probabilistic}.
% From a design philosophy perspective, both RAG and IR model ``relevance'' at multiple levels—including expression, meaning, relations, and usefulness to the final outcome.

We next focus on the retrieving module in RAG. This step concerns how queries are matched with evidence, motivating the study of matching mechanisms. In principle, this module can be instantiated with any existing retrieval method—ranging from classical algorithms such as BM25 to modern sparse and dense retrievers—since RAG treats retrieval largely as a pluggable component rather than prescribing a specific mechanism.

To improve retrieval precision, RAG has explored a wide spectrum of matching mechanisms. These range from \emph{sparse retrieval} that emphasizes lexical matching~\cite{robertson2009probabilistic,formal2021splade}, to \emph{dense retrieval} that captures semantic similarity in vector space~\cite{karpukhin2020dense,xiong2020approximate}, and further to \emph{graph- or tree-based retrieval} that models structured relations~\cite{sun2018open,sarthi2024raptor,edge2024local}. More recently, \emph{learning-based matching} functions have been introduced to align retrieval with the utility of downstream generation, optimizing not merely for topical relevance but for how much a retrieved passage actually improves the final generated answer.

Clearly, these developments closely mirror the evolution of traditional IR, where similar directions have been extensively studied~\cite{robertson2009probabilistic}: from term-based models, through vector-space and probabilistic representations, to learning-to-rank approaches that directly optimize for downstream retrieval effectiveness. From a design philosophy perspective, both RAG and IR model "relevance" at multiple levels—including expression, meaning, relations, and usefulness to the final outcome.

\begin{figure*}[t!]
    \centering
    % 之前的颜色
    % \includegraphics[width=\textwidth]{frameworks3-final-new.pdf}
    % 新颜色
    \includegraphics[width=\textwidth]{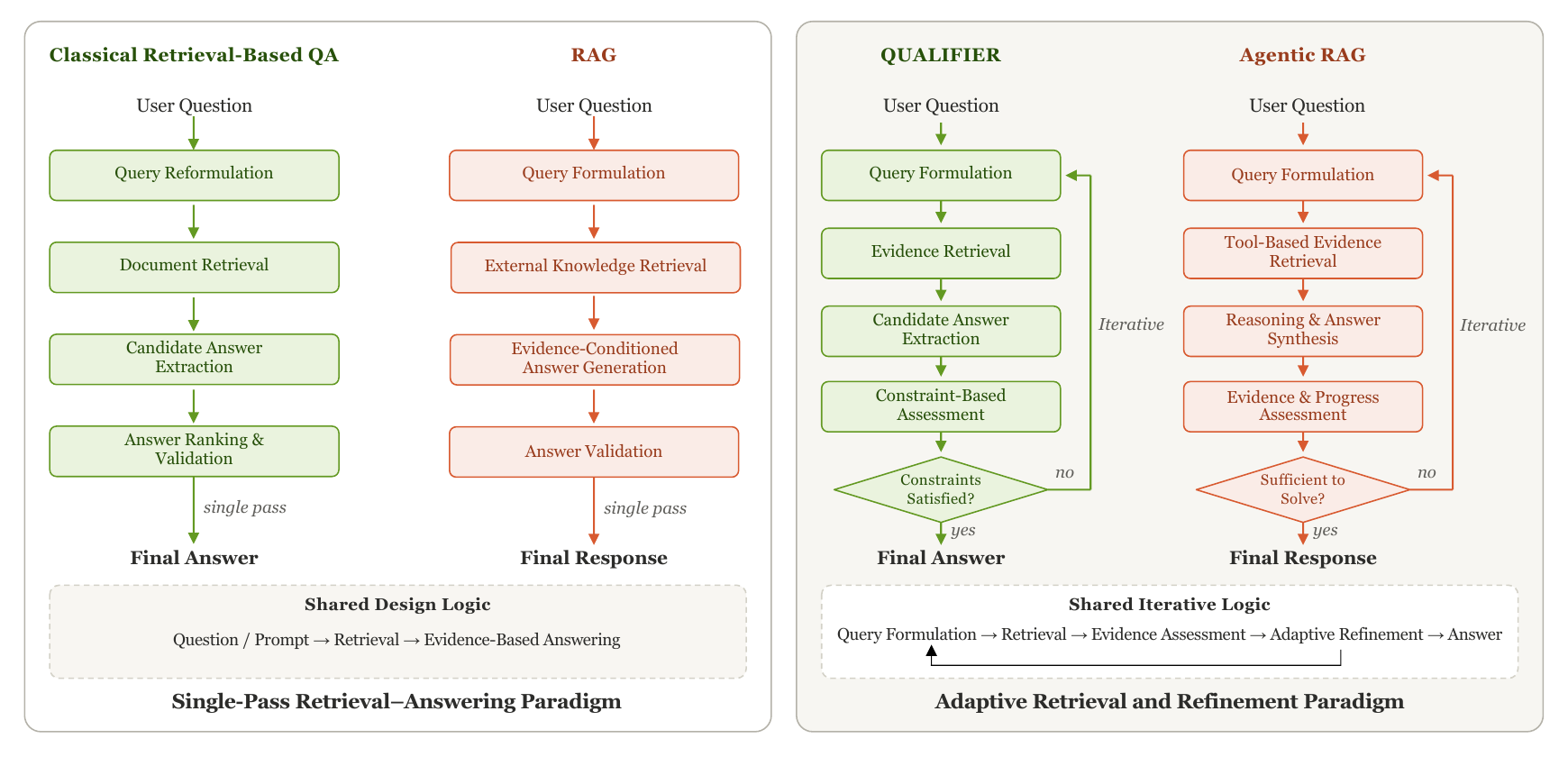}
    \caption{Conceptual continuity from classical QA to modern RAG. Classical QA and RAG share a single-pass retrieval–answering design, while QUALIFIER and Agentic RAG share an adaptive, iterative retrieval-and-refinement logic.}
    \label{fig:framework_compare} 
\end{figure*}

\subsection{Answer Synthesis: Evidence Access, Faithfulness, Quality Control}

Answer validation was already a central concern in early TREC-era QA systems, well before RAG formalized these ideas. PIQUANT \cite{ChuCarrollPWCF02} employed a dedicated sanity checker to filter implausible candidate answers before final selection. DIOGENE \cite{magnini2002mining} incorporated a complex answer validation module that assessed candidates against multiple internal criteria before returning a result. LCC \cite{Moldovan2001} took a more logic-driven approach, developing a logic prover that operated over heavily annotated logical forms stored in its customized WordNet. QUALIFIER \cite{yang2003structured} validated answers by requiring candidates to satisfy its strict structured-query constraints, admitting only those that passed this high bar. Most other systems, such as MultiText \cite{ClarkeCKLLTT02} and AskMSR \cite{brill2001data,brill2002analysis}, instead relied on web-based redundancy signals: an answer's correctness was estimated by how frequently it was corroborated across independently retrieved web snippets. Despite their differing mechanisms, all of these systems shared a common goal—verifying candidate answers before committing to them—that anticipates the quality-control concerns RAG systems face today.

In RAG, generation is better understood as an evidence synthesis process rather than unconstrained language generation. We organize its objectives along a progression ``evidence access → faithfulness → quality control'' that reflects the shift from merely using external knowledge to using it responsibly and reliably. Many mechanisms in RAG-based generation do not introduce fundamentally new design principles. Instead, they reinterpret ideas long explored in classical IR and question answering, particularly in evidence aggregation, attribution, and feedback-driven control.

%In RAG, generation is better understood as an evidence synthesis process rather than unconstrained language generation. We organize its objectives along a progression ``accessibility → faithfulness → quality control'' that reflects the shift from merely using external knowledge to using it responsibly and reliably. Many mechanisms in RAG-based generation do not introduce fundamentally new design principles. Instead, they reinterpret ideas long explored in classical IR and question answering, particularly in evidence aggregation, attribution, and feedback-driven control.

%\paragraph{Part 1: Accessibility} The accessibility provides the baseline contract of RAG: retrieved evidence must be accessible to condition generation so that knowledge is externalized beyond model parameters. This reflects the QA paradigm of retrieval-then-synthesis in classic IR researches. 

\paragraph{Part 1: Accessibility} Accessibility provides the baseline contract of RAG: retrieved evidence must be accessible to condition generation so that knowledge is externalized beyond model parameters. This reflects the QA paradigm of retrieval-then-synthesis in classic IR research.

\paragraph{Part 2: Faithfulness} In the agentic RAG system, answer generation is not simply text-conditioned writing but evidence-centered synthesis. Once accessibility is established, the critical question becomes not whether evidence is available, but whether the answer is derived from and traceable to that evidence. Generation should read multiple passages, selectively combine them, and produce answers that remain attributable to specific sources.
To achieve this, Agentic RAG adopts mechanisms such as fusion-style readers that aggregate information before writing~\cite{izacard2021leveraging}, and attribution-oriented training that links outputs to their supporting evidence~\cite{menick2022teaching,bohnet2022attributed}. These ideas are not new in spirit—they closely correspond to IR work on multi-document summarization and redundancy-aware diversification~\cite{carbonell1998use,radev2004centroid,lin2011class}, where systems must synthesize across many documents while preserving coverage and avoiding over-representation. The core philosophy shared by both is that answers must be justified by evidence.

\paragraph{Part 3: Quality Control} In real-world settings, retrieval evidence is inherently uncertain, and single-pass generation can be systematically biased by erroneous or incomplete evidence. Agentic RAG therefore reframes generation as a decision-centric process rather than a one-shot writer: the system could decide when to retrieve, how to assess retrieval quality, whether evidence should be refined or expanded, and when to abstain or revise, enabling explicit monitoring of outputs.
Recent RAG work instantiates this idea through self-reflection and retrieval adaptation~\cite{asai2024self}, corrective actions conditioned on retrieval-quality assessment~\cite{yan2024corrective}, and iterative attribution–revision workflows~\cite{gao2023rarr}, often regarded as core components of Agentic RAG. 

Figure~\ref{fig:framework_compare} summarizes these two levels of conceptual continuity: classical retrieval-based QA and modern RAG share an evidence-conditioned answering pipeline, while QUALIFIER and Agentic RAG further share an iterative retrieval and refinement loop.

%While classical IR did not directly employ modern agentic concepts, some underlying logic is continuous with the refinement ideas using relevance feedback~\cite{rocchio1971relevance}, pseudo-relevance feedback~\cite{lavrenko2001relevance}, or uncertainty. The common philosophy is adaptive quality control: results should be continuously evaluated and adjusted. In RAG, LLMs extend this idea with richer general capabilities.

\section{Proto-Agentic RAG: The QUALIFIER System}

%\subsubsection{Regime III: Addressing Reasoning Gaps via Constraint-Guided Inference}

% As QA tasks grew more complex, retrieving relevant documents alone proved insufficient for complex questions requiring synthesis, constraint satisfaction, and multi-step reasoning. This led to a shift from document-centric retrieval toward structured, constraint-guided reasoning answer construction.

% The QUALIFIER system~\cite{yang2003qualifier,yang2003structured} 
% modeled questions as structured semantic events composed of slots (e.g., time, location, agent, action), treating known components as constraints and inferring missing elements through reasoning over retrieved evidence. 
% leveraged external resources-including web snippets, WordNet \cite{moldovan2001logic}, and Wikipedia—to expand and validate candidate evidence under semantic constraints. When constraints were overly restrictive, successive constraint relaxation was applied to iteratively broaden the search space \cite{yang2003structured}. This introduced an early form of retrieval-guided reasoning and verification into QA, and it has the form of Agentic RAG. 

As QA tasks grew more complex, retrieving relevant documents alone proved insufficient for complex questions requiring synthesis, constraint satisfaction, and multi-step reasoning. This led to a shift from document-centric retrieval toward structured, constraint-guided reasoning answer construction.

The QUALIFIER system~\cite{yang2003qualifier,yang2003structured} 
modeled questions as structured semantic events composed of slots (e.g., time, location, agent, action), treating known components as constraints and inferring missing elements through reasoning over retrieved evidence. 
It leveraged external resources—including Web snippets and WordNet—to expand and validate candidate evidence under semantic constraints. When constraints were overly restrictive, successive constraint relaxation was applied to iteratively broaden the search space \cite{yang2003structured}. It introduced retrieval-guided reasoning and verification into QA, making it an early precursor to Agentic RAG.

\subsection{Iterative Query Refinement Driven by Answer Verification}

The dynamic view of modern agentic RAG asks how retrieval can be adapted throughout the process of problem solving. In this setting, retrieval is no longer a one-shot operation but an iterative process: evidence acquired so far conditions subsequent information acquisition and termination \cite{asai2024self, xiong2020answering, zhao2021multi}. This reframing closely mirrors IR's closed-loop paradigm, where relevance feedback and pseudo-relevance feedback iteratively update the query and refine results \cite{rocchio1971relevance, lavrenko2001relevance}.

In RAG, iterative retrieval typically manifests as (i) query refinement through rewriting, expansion, or disambiguation based on intermediate evidence, (ii) multi-hop acquisition for compositional questions, and (iii) controller-driven retrieval that allocates budget across rounds, with reranking and filtering serving as explicit control primitives \cite{glass2022re2g, yu2024rankrag}. These approaches are conceptually aligned with IR~\cite{liu2009learning,wang2011cascade}, while generative models introduce new considerations—such as latency–quality trade-offs and limited context windows—that reshape the practical design space.

The QUALIFIER system~\cite{yang2002integration,yang2003qualifier,yang2003structured,yang2003trec2003qualifier} evolved across three TREC participations, moving from simple query expansion toward a structured, event-based model of questions. In their first TREC QA paper, Yang and Chua~\cite{yang2002integration} expanded short factoid queries using terms co-occurring in top-ranked Web documents, re-weighted through WordNet glosses and synsets, then used the expanded query for Boolean retrieval over the TREC corpus via the MG indexing system~\cite{yang2002integration}. Because Boolean retrieval only returns documents matching all query terms, an overly expanded query could return no results at all. To handle this, they introduced successive constraint relaxation (SCR): when a query returned no exact answer, the system removed a portion of the expanded terms and re-ran retrieval and extraction against the smaller, less-restrictive query. This relax-and-retry cycle repeated for up to five iterations; if no exact answer was found after all five, the system returned NIL rather than a low-confidence guess, a design the authors described as helping "increase the recall while preserving precision."

In their follow-up SIGIR 2003 paper, Yang et al.~\cite{yang2003structured} extended this idea by reframing factoid questions as structured QA events-question elements such as time, location, subject, object, and action, mined from Web snippets, WordNet, and pre-retrieved TREC documents. Because these elements tend to co-occur in predictable patterns (e.g., a "discovery" event links an explorer, a location, and a time), the system mined association rules among event elements and used them to score and rank candidate passages via an Answer Event Score, rather than relying on flat term overlap alone~\cite{yang2003structured}. This gave constraint relaxation a more principled foundation: relaxation could now target specific underperforming event elements rather than only reducing raw term counts.

By their TREC 2003 (TREC-12) participation~\cite{yang2003trec2003qualifier}, the authors explicitly separated their evaluation runs by objective, submitting one run optimized purely for recall—using SCR aggressively with anaphora resolution and abbreviation co-reference but without answer justification—and another optimized for precision, where an answer justification module played the primary filtering role, with SCR serving only to keep recall at an acceptable floor rather than driving the search. Across all three iterations of the system, the constraint relaxation mechanism preserved the same underlying logic: begin retrieval and extraction under maximally strict constraints, and only relax them incrementally, and as a last resort, when the stricter configuration yields nothing—ensuring that recall gains are pursued only after precision-preserving options are exhausted.

Although implemented with symbolic representations and heuristic control, QUALIFER's underlying design philosophy—that retrieval should be iteratively shaped by intermediate reasoning states—closely parallels modern Agentic RAG architectures, where retrieval and generation interact through structured prompts, multi-step reasoning, and refinement mechanisms.

\subsection{Competitive Performance}

QUALIFIER, submitted under the run tag \textit{pris2002}, was one of the top-performing systems at TREC 2002, ranking just behind LCC's logic-based PowerAnswer system by number of correctly answered questions (290 of 500), and well ahead of systems submitted by strong IR teams including IBM, MIT, USC/ISI, BBN, and Waterloo~\cite{Voorhees2002QA}. Table \ref{tab:trec2002} lists the official TREC 2002 QA main task results for these systems, ranked by number of correctly answered questions. 

\begin{table}[h]
\centering 
\caption{TREC 2002 QA main task results by number of correctly answered questions (out of 500), adapted from Voorhees~\cite{Voorhees2002QA}.}
\begin{tabular}{lll}
\toprule
Run Tag & Participant & Answer Correctness \\
\midrule
LCCmain2002 & LCC (PowerAnswer) & 415 (83.0\%) \\
\textbf{pris2002} & \textbf{QUALIFIER (NUS)} & \textbf{290 (58.0\%)} \\
uwmtB3 & MultiText (Waterloo) & 184 (36.8\%) \\
IBMPQSQACYC & PIQUANT (IBM) & 179 (35.8\%) \\
aranea02a & Aranea (MIT) & 152 (30.4\%) \\
isi02 & USC/ISI & 149 (29.8\%) \\
BBN2002C & BBN & 142 (28.4\%) \\
\bottomrule
\end{tabular} \label{tab:trec2002}
\end{table}

At the time, these results seemed surprising, and little attention was paid to why a system built this way outperformed other mainstream QA systems. In hindsight, the explanation is clearer: QUALIFIER was, in effect, an early Agentic RAG system—its iterative query relaxation loop retrieves, checks whether an answer was found, and adaptively reformulates the query if not, rather than performing a single-shot retrieval-then-generate pass, as most other proto-RAG-ish QA systems did. This helps explain why it outperformed the single-shot, retrieval-then-generate QA systems—often reliant on predictive annotation—that dominated the field for the next decade, before community QA approaches emerged. QUALIFIER's competitive performance suggests that agentic RAG with iterative refinement can offer practical advantages over simple, single-pass RAG, even under benchmark conditions from over two decades ago.

\section{Discussion: What Is Missing Between IR and Modern RAG?}
As discussed previously, many components of modern RAG systems closely resemble ideas that have been explored in Information Retrieval for decades. Rather than emerging independently, these systems can be seen as continuing the same development trajectory in which similar challenges—such as lexical mismatch, knowledge incompleteness, and reasoning over evidence—are addressed under evolving technical regimes. What differs is not the underlying problem structure, but the modeling assumptions and implementation tools available at each stage.

One contributing factor to the under-recognition of this continuity lies in the divergence of research communities. Classical IR has focused on retrieval models, ranking functions, efficiency considerations, and user-centered analysis, whereas much RAG research focuses on LLM-related problems, where systems are typically framed around end-to-end generation quality. These distinct traditions influence how problems are posed and which system components are foregrounded. 
In addition, evaluation practices have evolved differently.  IR research has emphasized 
ranking-oriented 
user-centered evaluation, while RAG work primarily evaluates generated outputs using measures of faithfulness, relevance, or summarization quality.  
Moreover, the rapid pace of LLM-centered research has prioritized empirical progress and system integration, leaving limited opportunity to situate new architectures within a broader historical and methodological context. As a result, established IR principles often reappear in modern RAG systems without explicit acknowledgment of their lineage.

In this paper, we revisit the development of RAG from an IR perspective, viewing LLMs as a new interface layer that mediates between user intent and external knowledge. In this role, LLMs translate underspecified user queries into richer retrieval signals and synthesize retrieved information into coherent responses, while the underlying challenges of relevance, coverage, and uncertainty remain fundamentally retrieval-driven. This perspective allows us to place RAG within a more complete historical context and helps identify IR insights that remain underexplored in current RAG research. Building on this analysis, the next section (Section~\ref{sec: IR guides RAG}) illustrates how established IR perspectives can more systematically inform the future design of RAG systems.

% \begin{figure*}[h]
%     \centering 
%     % \includegraphics[width=\textwidth, trim={0cm 2cm 3cm 0cm}, clip]{section4-fig.pdf}
%     \includegraphics[width=0.9\textwidth]{sec4-fig2.pdf}
%     \caption{
%     IR informs four key directions for future RAG development: personalized RAG, proactive RAG, RAG with rigid governance, and user-centered evaluation.
%     }
%     \label{fig:sec4} 
% \end{figure*}

\begin{figure*}[h]
    \centering 
    \includegraphics[width=0.95\textwidth]{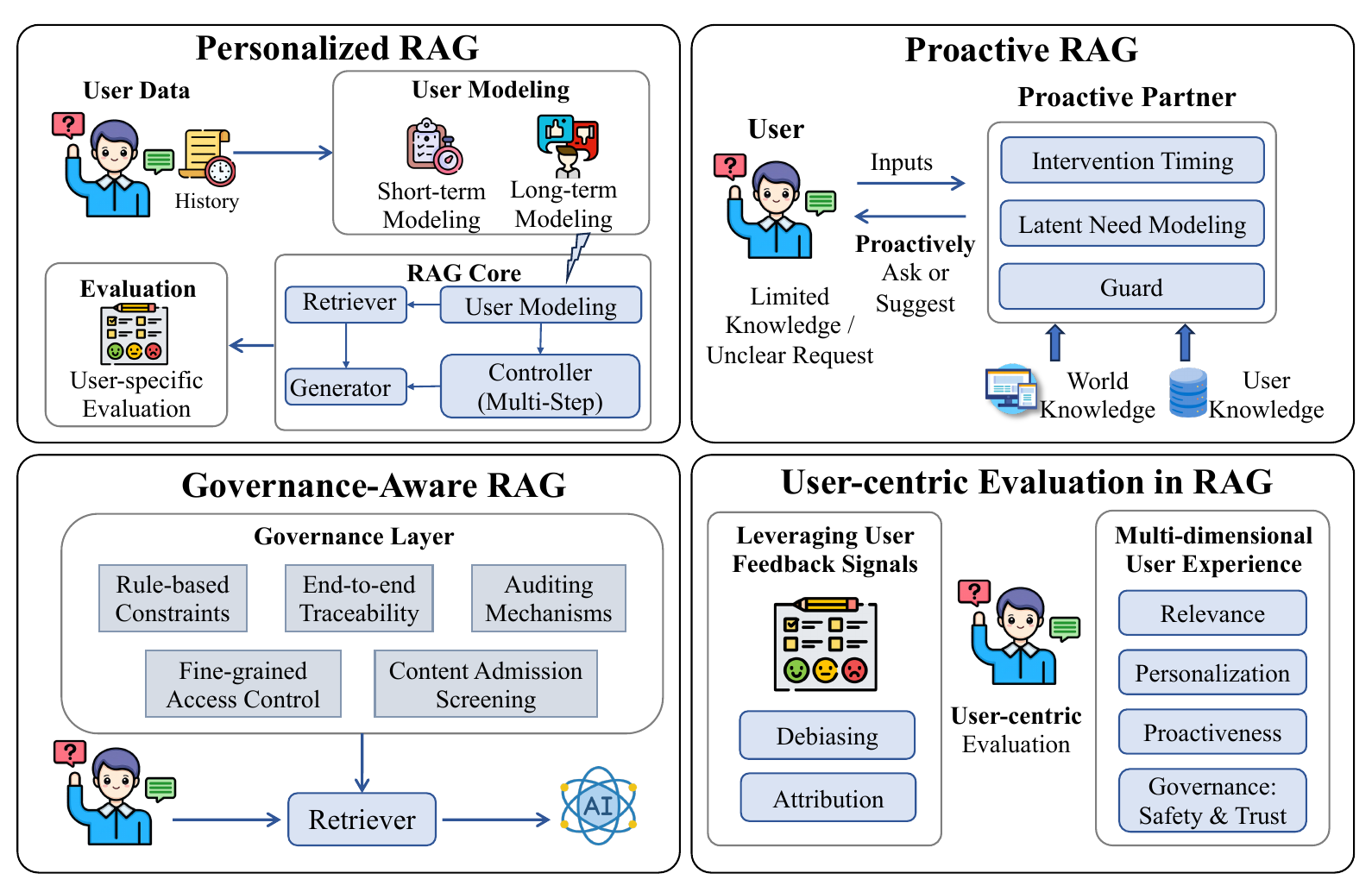}
    \caption{Four IR-informed directions for future RAG: personalization, proactive interaction, governance-aware architectures, and user-centric evaluation.
    }
    \label{fig:sec4} 
\end{figure*}

\section{Future Directions for RAG: Key Perspectives from IR}
\label{sec: IR guides RAG}

Although many RAG mechanisms can be understood as generative reinterpretations of long-standing IR principles, the IR literature contains a broader set of perspectives and methodological insights that remain underutilized in current RAG research. 
% We argue that the next stage of RAG should move beyond model-centric advances and revisit several IR-informed directions: personalized RAG, proactive RAG, Rigid Governance, and user-centered evaluation, as shown in Figure~\ref{fig:sec4}. 
\textcolor{black}{As the field moves beyond purely model-centric advances, several IR-informed directions have increasingly garnered attention, namely personalized RAG, proactive RAG, governance-aware RAG, and user-centered evaluation, as shown in Figure~\ref{fig:sec4}. We emphasize that these emerging areas will be crucial for the next stage of RAG development.}
These directions shift the focus from how to retrieve and synthesize information to for whom, when, and under what guarantees RAG systems should operate, thereby opening a broader design space for user-centered and accountable systems.

\subsection{Personalized RAG}

The first key perspective from IR for the future of RAG concerns personalization.

\subsubsection{IR Perspective 1: Personalization} 
In IR, it is emphasized that relevance is not a static notion of semantic match, but depends on the user’s evolving goals, task context, and interaction history. Interactive IR and “retrieval-in-context” research further argued that information access should be organized around the user situation, enabling systems to adapt to user needs and to reduce the cost of repeatedly articulating intent \cite{ingwersen1992information}. Building on this, later works pursued user-centered personalized retrieval services by constructing user models from behavioral signals, improving precision without imposing additional burden on users \cite{shen2005implicit}. Some efforts even aim to unify search and recommendation—leveraging rich user feedback from recommendation—to enhance user modeling for improved retrieval performance~\cite{ingwersen2005turn}.
Together, these developments reveal a core orientation of IR: to place the user at the center, achieving higher-quality and personalized information services while minimizing interaction and cognitive costs.

\subsubsection{Direction 1: Personalized RAG}
Current RAG systems largely operate at the level of isolated queries and have yet to fully incorporate personalization. However, when RAG is deployed in real-world scenarios—such as search, professional research, or decision support—personalization becomes essential for improving user experience and long-term utility.

\textcolor{black}{Recognizing this necessity, an emerging line of recent studies has begun to explore personalized RAG \cite{salemi2024optimization, zhang2025prlm}.} Personalized RAG shifts the optimization objective from single-query performance to user-level alignment, accounting for both short-term intent and long-term preferences. This requires integrating user modeling into the core retrieval–generation pipeline, enabling systems to adapt evidence selection, synthesis strategies, and control policies to individual users. Realizing Personalized RAG introduces several challenges:

% \vspace{-1em}
\begin{itemize}
[leftmargin=*]
    % \item[(i)]
    \item 
Better alignment with users requires informative user signals, yet many RAG scenarios provide only limited and noisy feedback. Moreover, these signals must guide long-form contexts and multi-step retrieval–generation processes, making reliable user learning particularly difficult. Future work may explore process-level reward assignment and detailed feedback simulation to provide richer supervision for personalization.

% \item[(ii)] 
\item 
Since user signals are sparse, RAG instances often involve long contexts, while the effective reasoning capacity of LLMs may degrade as context length increases. This tension complicates stable lifelong user modeling. Potential solutions include explicit memory modules and continual learning mechanisms that incrementally integrate user preferences into the generation process.

% \item[(iii)] 
\item 
Evaluating personalized RAG is another challenge because preferences are subjective and difficult to capture with existing metrics, particularly those based on semantic matching. Furthermore, RAG evaluation may extend beyond textual outputs to presentation and UI-level interactions, increasing complexity. Future work may need to shift the evaluation paradigm toward learning-to-evaluate for different users, generating adaptive meta-evaluation rules from diverse preference aspects rather than relying on fixed criteria.

\end{itemize}

\subsection{Proactive RAG}
\label{sec:proactive-rag}
The second perspective from IR for the future of RAG concerns proactive interaction.

\subsubsection{IR Perspective 2: Proactive Interaction}
In classical IR, user inputs are frequently incomplete, underspecified, or even difficult for users themselves to articulate due to uncertainty about their underlying needs. Recognizing this, IR systems introduced mechanisms for intent construction support, such as query auto-completion and query suggestion, which significantly improve user experience by reducing articulation cost and guiding need formulation~\cite{cai2016survey,bhatia2011query}.
This insight motivated the development of \textit{proactive interaction} in IR: systems began to engage users through clarification questions, facet elicitation, and suggestive prompts to surface latent intent and improve retrieval effectiveness~\cite{aliannejadi2019asking,zamani2020generating,zamani2020analyzing}.
From this perspective, RAG systems possess similar potential: beyond responding to explicit queries, they can assist users in refining, structuring, and even discovering their underlying information needs.

\subsubsection{Direction 2: Proactive RAG}
RAG systems remain largely dependent on the accuracy of user-provided queries and continue to operate in a predominantly passive interaction paradigm, waiting for users to articulate their needs. In this sense, they inherit many of the same limitations as traditional IR systems. We argue that future RAG should move beyond this passive framework and incorporate proactive interaction as a core capability, forming proactive RAG. 
This is particularly important for RAG systems designed for real-world user interaction: users usually vary widely in background, expertise, and task familiarity, and their ability to precisely formulate information needs differs across topics. Proactive assistance can therefore play a crucial role in helping users better articulate the problems they aim to solve.
By proactively helping users finish tasks, proactive RAG has the potential to substantially enhance the overall user experience and transform RAG from a passive answer engine into a collaborative problem-solving partner.

To realize Proactive RAG, we face challenges similar to those encountered in proactive IR. First, intervention timing is crucial: while proactive actions may reduce uncertainty, they also incur interruption costs. Systems must therefore balance expected utility against user burden. Second, latent need modeling lies at the heart of effective interaction. A central difficulty is how to surface and refine users’ underlying needs. Third, proactive intervention must preserve user autonomy, avoiding excessive intrusion while still providing meaningful guidance. 

These challenges are amplified in RAG systems. Unlike classical IR, where proactive actions primarily influence query formulation, in RAG, they directly affect downstream reasoning and generation. Therefore, intervention decisions shape not only retrieval but also how evidence is interpreted and synthesized.
At the same time, the integration of LLMs and agent-based reasoning offers new opportunities. With stronger world knowledge and reasoning capabilities, RAG systems may better infer latent intent and make more principled intervention decisions. This suggests that Proactive RAG represents both a continuation of proactive IR principles and an expanded design space enabled by generative models.

\subsection{Governance-Aware RAG}
In this section, we discuss lessons in IR from the perspectives of safety and trustworthiness, providing insights into the future development of RAG.

\subsubsection{IR Perspective 3: Structured Governance}
 In the evolution of IR, it became clear that system success depends not only on accuracy but also on trustworthiness. As a primary channel for information access, IR systems profoundly influence individuals and, at scale, society. Consequently, trust, compliance, and social responsibility emerged as central design concerns. IR research has increasingly focused on safety and trust, addressing issues such as bias, fairness, and regulatory compliance, recognizing that reliability extends beyond retrieval effectiveness~\cite{olteanu2021facts,morik2020controlling,zerveas2022mitigating}.
As a mature and widely deployed real-world technology, IR has developed mature governance mechanisms. Through rule-based filtering, fine-grained access control, authoritative source screening, and end-to-end auditing~\cite{kleinberg1999authoritative,sayed2019jointly}, IR systems have established enforceable safeguards over pipelines.

\subsubsection{Direction 3: Governance-Aware RAG}
For RAG systems, safety and trustworthiness are also critical. Recent work has explored safety and trust issues~\cite{wallat2025correctness}. However, current RAG approaches often concentrate on input and output filtering at the language model level, and have not yet developed governance mechanisms that are as systematically integrated into the full retrieval–generation pipeline as in mature IR systems.

A promising direction is to incorporate structured governance principles from IR into both the retrieval and generation layers of RAG.  This includes integrating strict retrieval guardrails, rule-based constraints, fine-grained access control, content admission screening, and end-to-end traceability and auditing mechanisms~\cite{gill2025search,xu2025ragops,zou2025poisonedrag}. However, enforcing rigorous governance in RAG is more challenging than in traditional IR systems. Unlike conventional IR, RAG inherently incorporates the LLM generation process, which is characterized by high uncertainty~\cite{wallat2025correctness}. Implementing strict governance under such uncertainty undoubtedly raises greater technical hurdles, but it also opens up valuable opportunities.

\subsection{User-Centered Evaluation in RAG}
\subsubsection{IR Perspective 4: User-Centered Evaluation}

Classical IR has developed an evaluation discipline that extends well beyond static relevance judgments. A key advancement lies in the refined interpretation of user feedback. Rather than treating clicks or interaction signals as direct indicators of relevance, IR research accounts for systematic biases such as position and exposure effects, enabling more accurate evaluation of system quality~\cite{joachims2002optimizing,craswell2008experimental}. Beyond real-time feedback, IR also incorporates controlled user studies to assess subjective and long-term user experience, including perceived usefulness and satisfaction~\cite{chapelle2009dynamic}. In this way, IR evaluation moves beyond surface-level relevance and adopts a user-centered perspective that considers how users interpret, experience, and benefit from retrieval systems~\cite{kelly2009methods}.

\subsubsection{Direction 4: User-centered Evaluation in RAG}
In RAG systems, evaluation plays a dual role: it assesses methodological effectiveness and simultaneously provides learning signals that guide model alignment. Given the generative nature of RAG—particularly in open-ended generation tasks—user-centered evaluation becomes both more essential and more challenging. 
From a methodological perspective, user-centered evaluation more directly reflects real user satisfaction than static reference-based metrics. From a learning perspective, authentic user feedback offers clearer guidance for aligning model behavior with user preferences and long-term utility. Although existing RAG work has incorporated human evaluation, such assessments are typically conducted by annotators and may not faithfully represent real user experiences, as they often lack users’ historical context and preference backgrounds.

A promising direction, therefore, is to more fully and precisely leverage authentic user feedback to enable genuinely user-centered evaluation. 
This involves, on the one hand, carefully modeling and learning implicit user feedback signals, and on the other, evaluating user experience across multiple dimensions beyond factual correctness—for example, assessing personalization, proactiveness, and governance, as discussed in the first three directions.

Compared with classic IR, user-centered evaluation in RAG introduces additional complexity. Unlike traditional retrieval or extractive QA, RAG outputs are generated by LLMs and may not strictly correspond to retrieved evidence, complicating grounding assessment and increasing process uncertainty. Moreover, the RAG pipeline spans query reformulation, retrieval, reasoning, and generation, making it difficult to attribute user satisfaction or dissatisfaction to specific components. Finally, generative variability means that the same query may yield heterogeneous responses across users or runs, further complicating consistent evaluation.
These characteristics make user-centered evaluation in RAG both more necessary and structurally more challenging than in classical IR, highlighting the need for new evaluation paradigms.

\section{Related Work}

Existing surveys have examined RAG and IR from complementary perspectives. RAG surveys primarily characterize contemporary architectures and applications. Gao et al.~\cite{gao2023retrieval} categorize RAG into Naive, Advanced, and Modular paradigms, Zhu et al.~\cite{zhu2023large} review its development across modalities and downstream tasks, and Singh et al.~\cite{singh2025agentic} focus on agentic workflows involving planning, reasoning, and tool use. These studies generally take the post-LLM era as their starting point and frame retrieval as a mechanism for addressing limitations of parametric models, such as outdated knowledge and hallucination.

Meanwhile, IR surveys document the evolution from lexical retrieval to neural and semantic models~\cite{hambarde2023information}, examine the roles of LLMs in query rewriting, retrieval, reranking, and reading~\cite{zhu2023large}, or study generative retrieval methods that directly produce document identifiers and rankings. However, these lines of work rarely examine the historical continuity between classical retrieval-based QA and modern RAG. In contrast, we trace shared principles—including query reformulation, knowledge augmentation, evidence synthesis, verification, and iterative refinement—across these technical regimes, and further discuss how established IR perspectives can inform future RAG design, evaluation, and governance.

\section{Conclusions}
The development of Information Retrieval has long centered on connecting users’ information needs with reliable, relevant, and usable knowledge. From this perspective, RAG can be understood as a modern reconfiguration of long-standing IR and QA principles, including relevance modeling, evidence aggregation, iterative retrieval, constraint-guided reasoning, answer validation, and user modeling. Although these principles are now implemented through neural representations and generative control, their underlying design logic has been explored across decades of retrieval and question answering research. Recognizing this continuity is more than a historical observation: it helps clarify design trade-offs, avoid repeated rediscovery, and identify where accumulated IR experience can inform modern RAG systems.

Looking forward, the next stage of RAG extends beyond retrieval accuracy and generative fluency. As discussed in Section~5, it requires stronger integration of personalization grounded in user modeling, proactive interaction that helps users refine latent needs, governance-aware architectures with enforceable constraints and traceability, and user-centered evaluation based on authentic feedback and bias-aware metrics. Translating these established IR concerns into generative and agentic settings will be essential for building RAG systems that are more reliable, adaptive, accountable, and aligned with real user needs.

\bibliographystyle{ACM-Reference-Format}
\bibliography{references}

\end{document}